# A Fast Fractal Image Compression Algorithm Using Predefined Values for Contrast Scaling

H. Miar Naimi, M. Salarian
Electrical Engineering Faculty
University Of Mazandran
Tel-Fax: 98-111-3239214
***Email: h_miare@nit.ac.ir , M_Salarian@nit.ac.ir***

**Abstract**
In this paper a new fractal image compression algorithm is proposed in which the time of encoding process is considerably reduced. The algorithm exploits a domain pool reduction approach, along with using innovative predefined values for contrast scaling factor, ***S***, instead of scanning the parameter space [0,1]. Within this approach only domain blocks with entropies greater than a threshold are considered. As a novel point, it is assumed that in each step of the encoding process, the domain block with small enough distance shall be found only for the range blocks with low activity (equivalently low entropy). This novel point is used to find reasonable estimations of ***S,*** and use them in the encoding process as predefined values, mentioned above. The algorithm has been examined for some well-known images. This result shows that our proposed algorithm considerably reduces the encoding time producing images that are approximately the same in quality.

***Keywords: Image compression, fractal coding and multi resolution.***

## 1. Introduction

Fractal image compression is widely used in image processing applications such as image signature [1], texture segmentation [2], feature extraction [3], image retrievals [4,5] and MR, ECG image processing [6]. However, this method suffers from a long encoding time as its main drawback. This long encoding time arise from very large number of domain blocks that must be examined to match each range block. The number of range blocks with size of $n \times n$, in an $N \times N$ image, is $(N/n)^2$, while the number of domain blocks is $(N-2n+1)^2$. Consequently it can easily be shown that the computation for matching range blocks and domain blocks has complexity of $O(N^4)$ [7]. Thus reducing this encoding time is a focus of research with practical ramifications. Several methods have been proposed to overcome this problem. One common way is the classification of blocks in a number of distinct sets where range and domain blocks of the same set are selected for matching. Here, the encoding time is saved at cost of image quality. Reducing the size of domain pool is another method that has been employed in several manners. In some approach domain blocks with small variance [3] and in some others domain blocks with small entropies were deleted from the domain pool [7]

In addition to the size of the domain pool, the computational cost of matching a range block and a domain block has an important role in encoding time. We reduced this cost by estimating the approximate optimum values for contrast scaling factor, ***S,*** instead of searching for it. Combining these two novel points, we propose a new fractal image coding that has a considerable shorter encoding time than the next fastest algorithm [7]. In section 2 we present a brief description of the fractal image coding. The proposed algorithm is presented in section 3. In section 4 the methodology and the results are presented and compared with the next fastest algorithm. Finally, in section 5 conclusions are presented and some future works are addressed.

## 2. Fractal Image Coding: A Brief Review

At the first step in fractal coding an image is partitioned into none overlapping range blocks of size $B \times B$, where $B$ is a predefined parameter [4,5,8]. Then a set of domain blocks are created from the original image, taking all square blocks of size $2B \times 2B$ with integer step L, in horizontal and vertical directions. Within each member in the domain pool, three new domain blocks are created by clockwise rotating it 90º, 180º and 270º, also these three and the original domain block all are mirrored. Here, in addition to the original domain block, we have seven new domain blocks. These blocks are added to the domain pool. After constructing domain pools (related to each range block) we must select the best domain block from domain pool and find an affine transformation that maps the selected domain block with minimum distance. The mentioned distance between a range block, $R$, and a decimated domain block, $D$, both with $n$ pixels is defined as follows:

$$E(R,D) = \sum_{i=1}^{n} (sd_i + o - r_i)^2 \qquad (1)$$

The best coefficient ***S*** and ***O*** are [9]:

$$s = \frac{< R - \overline{R}.1, D - \overline{D}.1 >}{\| D - \overline{D}.1 \|^2} \quad (2)$$

$$o = \overline{R} - s\overline{D} \quad (3)$$

$<,>$, *R, D,* $\overline{R}$ and $\overline{D}$ are inner product, range block, domain block, mean of *R* and mean of *D* respectively. Because of high computational cost of (2), it is convenient to search $s$ across a pre-sampled set of [0,1], instead of calculating (2).

Along the matching process, the best found transformations are only saved for range blocks which have been mapped with an acceptable error. The remaining range blocks are split into 4 new smaller range blocks, and the matching process is restarted for the new set. For example, if range blocks initially have a size of $16 \times 16$ pixels, the range blocks of the succeeding steps will have a size of $8 \times 8$, $4 \times 4$ and $2 \times 2$ respectively, that leaves a four step algorithm. Two strategies were used to reduce the encoding time in fractal coding algorithms. In his research, Saupe found that the domain pool is not necessary to include all of possible domain blocks and only the high variance blocks are sufficient [3,10]. In another work, the entropy measure was used instead of variance [7]. This entropy based method is superior to the Saupe method so we compare our algorithm to the entropy based one.

## 3. The proposed algorithm

In this paper we use two novel points to reduce the encoding time. The first point is restricting the domain pool to high entropy domain blocks. This causes the total evaluation time for finding related domain block of a range block to become shorter. The entropy of a block is defined below. Suppose $N$ be a $K \times K$ block of an Image as shown in figure 1

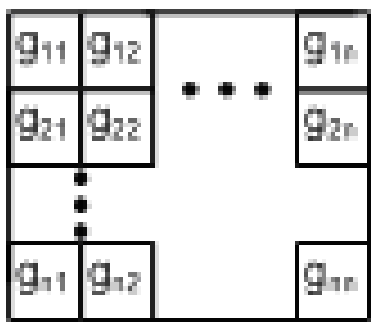

**Figure 1, a domain block of size** $K \times K$

In the above figure $g_{ij}$ is the grey level of the pixel at location $(i, j)$. Suppose $g_{ij}$ for $i, j = 1, 2, ..., n$ varies in $\{L_1, L_2, ..., L_K\}$. Also suppose the number of observations of $L_i$ over the pixels is $q_i$. So the probability of $L_i$ is defined as equation 5,

$$p_i = \frac{q_i}{\sum_{j=1}^{K} q_j} = \frac{q_i}{n^2} \quad (4)$$

The entropy is defined as below:

$$entropy = -\sum_{i=1}^{K} p_i Ln(p_i) \quad (5)$$

Here is some example block with size of $32 \times 32$

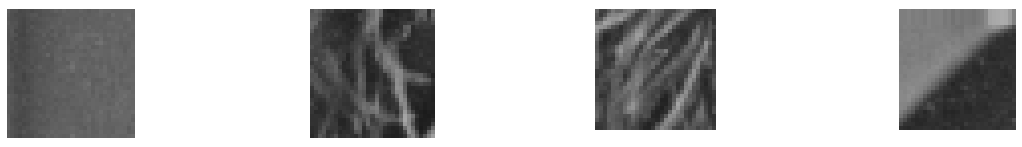

35.7 44.20 46.09 40.57

**Figure 2 four domain blocks and their related entropy**

As evident from figure 2, low entropy blocks are smoother and consequently have lower information contents. Lacking high frequency information, low entropy blocks cannot cover high entropy range blocks. On the other hand, high entropy blocks may cover all range blocks. To cover low entropy rang blocks we can simply reduce information of the domain blocks.

### 3.1 The effect of contrast scaling factor, $s$

Another important parameter that was investigated is the contrast scaling factor $s$. To do this, a large number of experiments with exhaustive search for $s$ were performed. Histograms of the best selected values of $s$ are shown in figure 3 for all four steps respectively. To analysis the effect of $s$, it will be helpful to recall the operation of $s$. As mentioned in section 1, domain block pixels are multiplied by $s$ and then the integer part is considered. Indeed, $s$ maps integer values of domain pixels to integer values of range pixels $(0 < s < 1)$. Here a simple proposition is presented that helps us interpret the presented histograms in figure 3.

**Proposition**

Suppose $0 < s_1 < s_2 < 1$ and $X_1, X_2$ and $Y$ are three sets of positive integer values with the same size. If $X_1 = [s_1 Y]$ and $X_2 = [s_2 Y]$ then

$$Entropy(X_1) \le Entropy(X_2)$$

* $[x]$ is the biggest integer less than $x$.

**Proof**: There is a simple proof as follows

$$0 < s_1 < s_2 < 1$$

$$0 < X_1 = [s_1 Y] \text{ , } 0 < X_2 = [s_2 Y]$$

$$\Rightarrow$$

$$Max(X_1) - Min(X_1) < Max(X_2) - Min(X_2) < Max(Y) - Min(Y)$$

(Here $0 < s < 1$ has a contractive role).

On the other hand we have

$$N(X_1) = N(X_2) = N(Y)$$

Here *N(X)* is the size of the set X. Thus X1 and X2 have the same size, but with different domain of variation and also are processed from the same set through a simple multiplication. It is obvious that the redundancies of $X_1$ will be larger than the ones of $X_2$. This simply concludes the results. At step 1 range blocks are $16 \times 16$ or of size 256 pixels. Consider now a block with a determined entropy or information. It is obvious that all permutations constructed by rearranging pixels of the block have the same entropy as the original. A simple and qualitative measure or as a lower bound for the number of these permutations is as follows:

$$N_P = \frac{256!}{n_1!n_2!\cdots n_k!} \qquad (6)$$

where $n_j$ is the number of pixels with grey level of j, big $n_j$ means the block has more redundancies and equivalently low entropy. Large values for $n_j$ are indicative of small distinct permutations. As a result, at step 1 only range block with small entropy will have the chance to be coded and consequently s has small values (recall the proposition above). If the entropy of a block is high at step one, then the number of blocks with that entropy will be high ($n_j$ s are small) so the probability that it could not be coded at this step would be high. Therefore, we expect that $s$ would have a small value. At lower steps 2, 3 and 4 block sizes are $8\times 8$, $4\times 4$. Following the same logic we see that $N_P$ drastically decreases. As a qualitative comparison we write:

$$\frac{NP2}{NP1} \propto \frac{64!}{256!} \approx 0$$

Again, with a similar discussion it may be shown that, blocks of higher entropy at level 2 are encoded so $s$ is left at greater values. This will also happen in lower steps. The histogram of the best $s$ in lower steps, according above discussion, will be shifted to the right, as shown in figure 3. In each step of existing algorithms, all members of a 10-member set of $s$, sampled from [0, 1], are evaluated. It can be seen from figure 3 that all values of $s$ need not be evaluated and we can restrict $s$ to one or two distinct values.

Obviously, restricting the size of $s$ to a 2-member set will decrease the search and hence the encoding time considerably. To find a true estimation of $s$, a large number of experiments with an exhaustive search for $s$ were performed. One can easily see that at step 1 the optimal $s$ value is often less than 0.1, independent of the image, so for this step we may let $s = 0.1$. At step 2 the optimum value of S is less than 0.5 so here we choose $s$ to be $\{0.2, 0.4\}$. For step 3, $s$ has approximately a uniform distribution across $[0,1]$, so to determine some distinct values here we choose S from {0.3, 0.8}. For step 4 as shown in figure 3d, $s$ the higher value in [0 1]. Here $s$ is chosen from $\{0.5, 0.9\}$. In this step

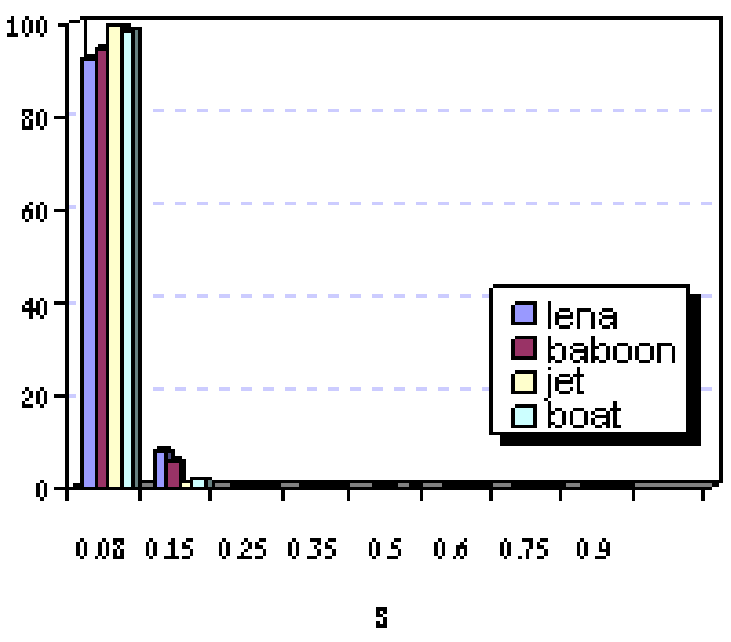

**(3a)**

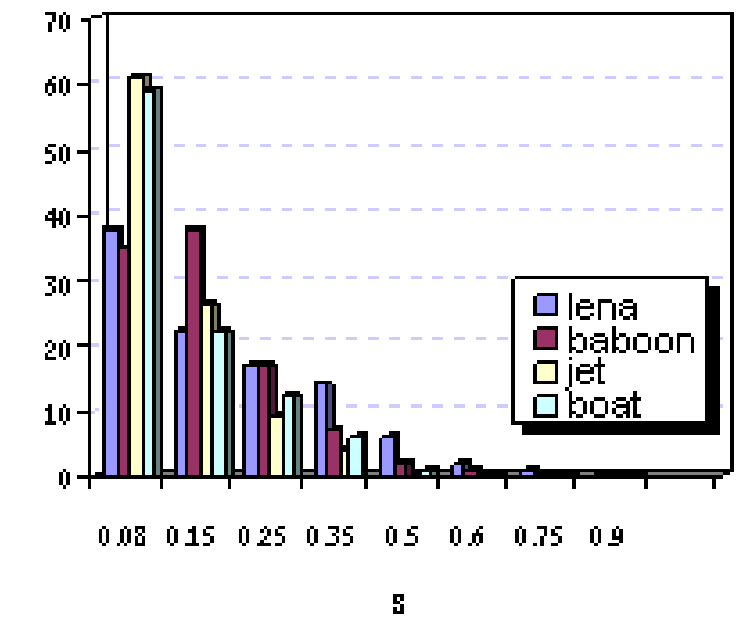

**(3b)**

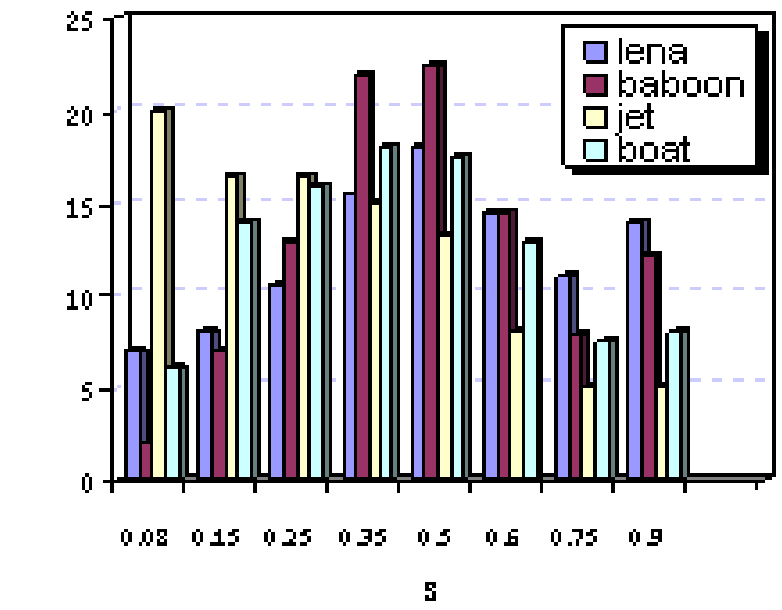

**(3c)**

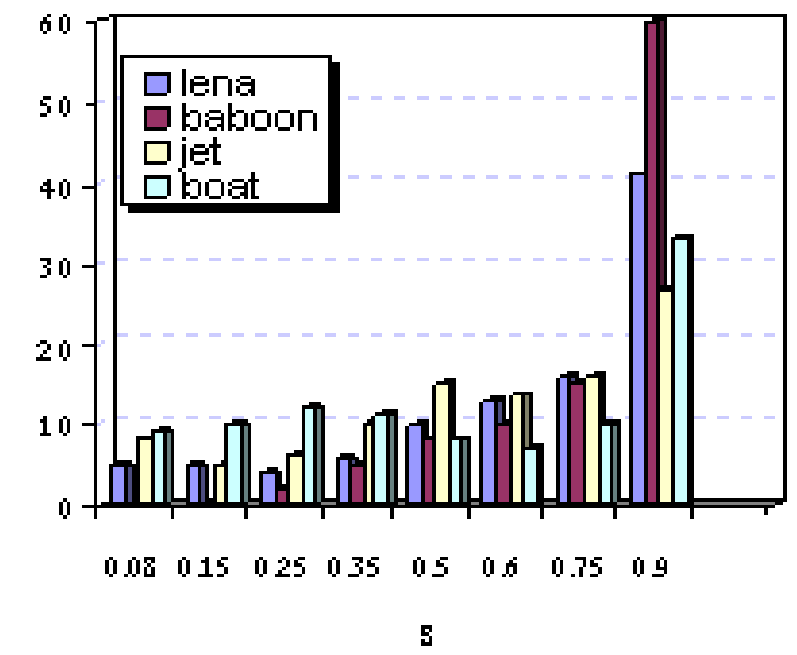

**(3d)**

***Figure 3 Histogram of $S$ at a) step1 b) step2 c)step 3 d)step 4***

blocks' size are $2\times 2$ that cause to be encoded very well. We reduced the set of values of $s$ to a two-member set that leaves three cases for range blocks. The first case is where the selected value is the same value as obtained from the exhaustive search of $s$ here isn't any problem. The second case is where the selected $s$ is not the best value, but the error is less than the threshold and the range blocks are coded approximately optimal here the encoding time is saved but the PSNR is somewhat damaged.
In the third case, the selected value causes the encoding error to become so large that range blocks can not be encoded. Hence, the range blocks are split and the encoding is done in following steps. This means a better PSNR at the cost of small degradation of time and compression ratio.

## 4. Experiments and results

Several experiments were performed to evaluate the proposed algorithm and compare it with the existing entropy based methods. Computer programs employed in these experiments were written in C++ running on a Pentium 2 (450MHz) with 256 MB RAM. Comparison results are shown in figure 4a, b, for different pool sizes and the Lena image. To have a reasonable comparison, the two algorithms are compared in fixed PSNR. Figure 4a,b show the compression ratio and encoding time for PSNR=35.07db. In these figures compression ratio and encoding time are plotted versus pool size with the PSNR as the parameter.

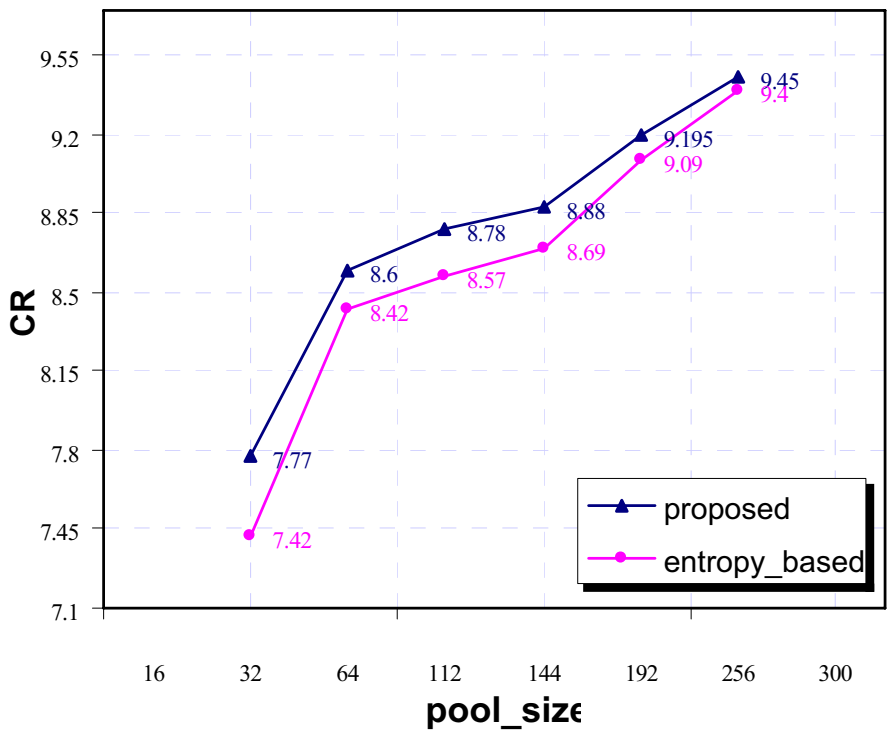

**(4a)**

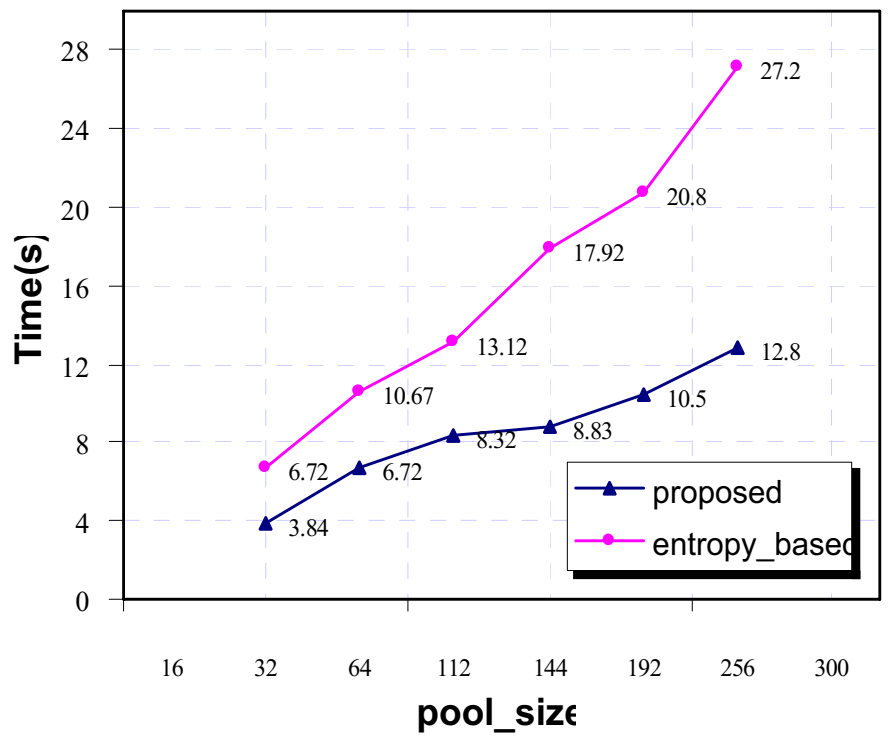

**(4b)**

Figure 4, Compression ratio and encoding time of proposed algorithm and entropy based versus pool size a, b) at fixed PSNR=35.07d***b***

To gain a greater perception of proposed algorithm the results for two other familiar images are presented in tables 1and 2. Comparing the two algorithms, it is evident that the proposed algorithm is superior especially in encoding time economy. The results of proposed algorithm for Lena image are presented in figure 6.

***Table 1 the comparison results for Baboon***

| | Prop | | | Entropy | | |
|---|---|---|---|---|---|---|
| Pool size | Com. rat | Enc. Time (S) | PSNR | Com. rat | Enc. Time | PSNR |
| 256 | 5.36 | 26.24 | 26.33 | 5.35 | 47.04 | 26.34 |
| 64 | 4.98 | 10.88 | 26.40 | 4.84 | 18.56 | 26.09 |
| 32 | 4.62 | 7.68 | 26.07 | 4.61 | 12.16 | 26.07 |

***Table 2 the comparison results for F16***

| | Proposed | | | Entropy based | | |
|---|---|---|---|---|---|---|
| Pool size | Com. rat | Enc. Time (S) | PSNR | Com. rat | Enc. Time | PSNR |
| 256 | 11.25 | 13.76 | 33.41 | 11.50 | 85 | 33.41 |
| 64 | 9.63 | 5.44 | 33.87 | 9.47 | 25 | 33.97 |
| 32 | 9.66 | 4.16 | 33.65 | 9.5 | 21 | 33.64 |

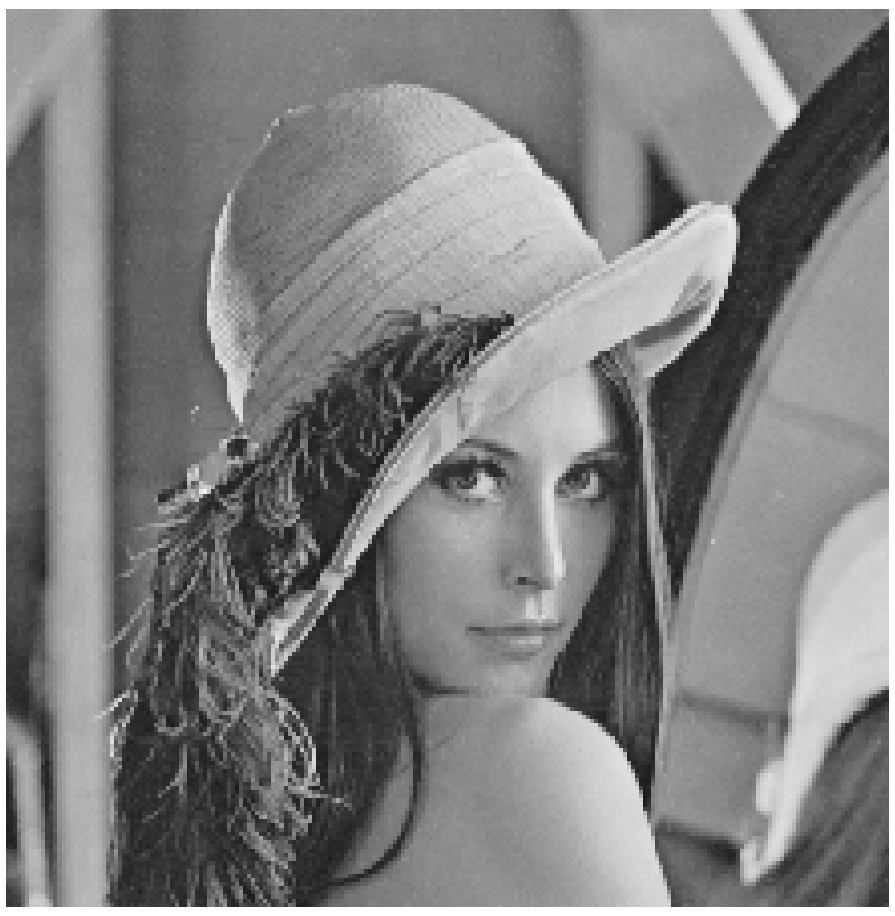

**Figure 5 Original Image**

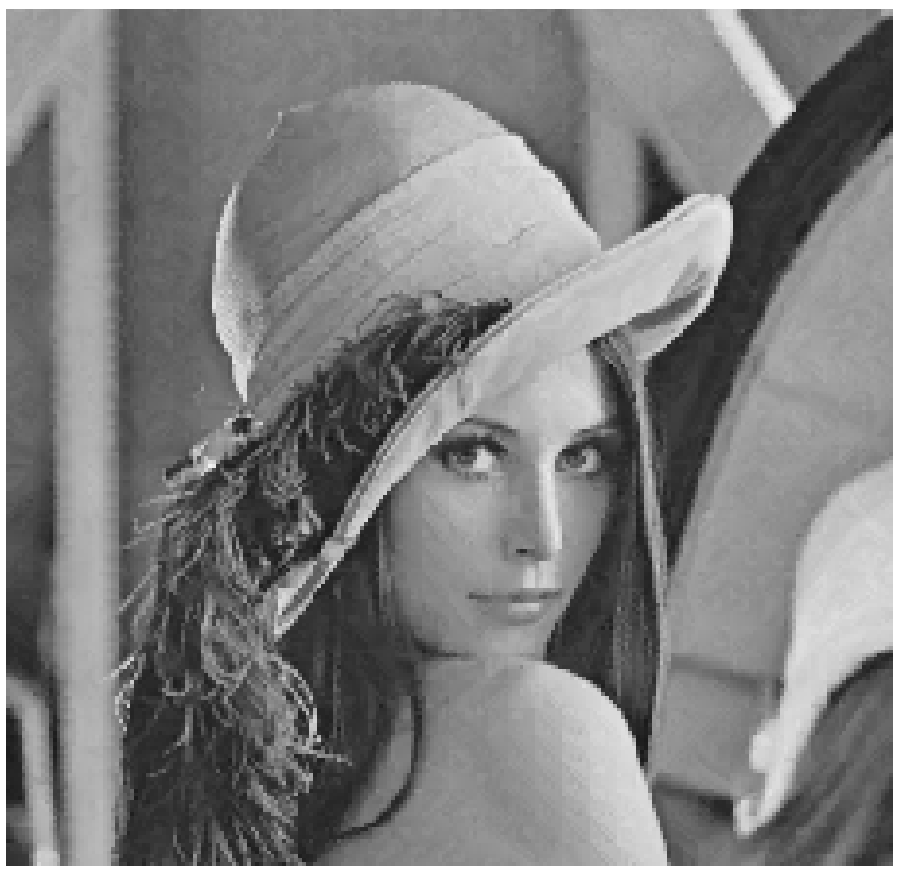

**Figure 6 Com.Rat=12.17 Time(8.2 s)**

**PSNR=33.57db**

## 5. Conclusions and future works

In this paper we presented a new method for fractal image compression to reduce encoding time. Centrally, our algorithm employed predefined values for contrast scaling factor rather than sweeping the entire parameter space during search. Experimental results indicate a superior performance level in comparison to the existing entropy based methods. In the future we intend to further develop this approach in frequency domain applications and produce quantitative comparisons with other hybrid methods.